\documentclass[letterpaper]{article} 
\usepackage{aaai23}  
\usepackage{times}  
\usepackage{helvet}  
\usepackage{courier}  
\usepackage[hyphens]{url}  
\usepackage{graphicx} 
\usepackage{natbib}  
\usepackage{caption} 
\usepackage{algorithm}
\usepackage{algorithmic}

\usepackage{graphicx}
\usepackage{amsmath}
\usepackage{bm}
\usepackage{multirow}
\usepackage{enumitem}
\usepackage{amssymb}
\usepackage[switch]{lineno} 
\usepackage{multirow}
\usepackage{threeparttable}
\usepackage{booktabs}

\usepackage{newfloat}
\usepackage{listings}
\DeclareCaptionStyle{ruled}{labelfont=normalfont,labelsep=colon,strut=off} 
\floatstyle{ruled}
\newfloat{listing}{tb}{lst}{}
\floatname{listing}{Listing}
\title{BEST: BERT Pre-Training for Sign Language Recognition \\ with Coupling Tokenization}
\author{
    Weichao Zhao{\small $~^{1}$}, Hezhen Hu{\small $~^{1}$}\thanks{Contribute equally with the first author.}, Wengang Zhou{\small $~^{1,2}$}\thanks{Corresponding authors: Wengang Zhou and Houqiang Li.}, Jiaxin Shi{\small $~^{3}$}, Houqiang Li{\small $~^{1,2\dag}$}
}
\affiliations{
    \textsuperscript{\rm 1}CAS Key Laboratory of GIPAS, EEIS Department, University of Science and Technology of China (USTC) \\
    \textsuperscript{\rm 2}Institute of Artificial Intelligence, Hefei Comprehensive National Science Center 
    \textsuperscript{\rm 3}Huawei Inc. \\
    \normalsize
{\tt\small \{saruka, alexhu\}@mail.ustc.edu.cn, \{zhwg, lihq\}@ustc.edu.cn, shijiaxin3@huawei.com}
}

\usepackage{bibentry}

\begin{document}

\maketitle
\begin{abstract}
In this work, we are dedicated to leveraging the BERT pre-training success and modeling the domain-specific statistics to fertilize the sign language recognition~(SLR) model.
Considering the dominance of hand and body in sign language expression, we organize them as pose triplet units and feed them into the Transformer backbone in a frame-wise manner.
Pre-training is performed via reconstructing the masked triplet unit from the corrupted input sequence, which learns the hierarchical correlation context cues among internal and external triplet units.
Notably, different from the highly semantic word token in BERT, the pose unit is a low-level signal originally located in continuous space, which prevents the direct adoption of the BERT cross-entropy objective.
To this end, we bridge this semantic gap via coupling tokenization of the triplet unit.
It adaptively extracts the discrete pseudo label from the pose triplet unit, which represents the semantic gesture/body state.
After pre-training, we fine-tune the pre-trained encoder on the downstream SLR task, jointly with the newly added task-specific layer.
Extensive experiments are conducted to validate the effectiveness of our proposed method, achieving new state-of-the-art performance on all four benchmarks with a notable gain.

\end{abstract}

\section{Introduction}
Sign language is a primary communication tool for the deaf community.
It is characterized by its unique grammar and lexicon, which are difficult to understand for non-sign language users.
To bridge this communication gap, automatic sign language recognition~(SLR) is widely studied with broad social influence.
As one basic task, isolated SLR aims to recognize at the gloss-level and is a fine-grained classification problem. 
In this work, we focus on this task.

Due to high annotation cost, current labeled sign data sources are limited.
Since common deep-learning-based methods are data-hungry, they are prone to over-fitting on SLR. To this end, several attempts have been made in SLR. 
For instance, considering the dominant role of hand, some methods~\cite{hu2021hand, albanie2020bsl} utilize the cropped hand sequence as the auxiliary information.
HMA~\cite{hu2021hand} proposes to recognize sign language in a model-aware paradigm with the hand mesh as the intermediate constraint.
However, those methods directly optimize on the target benchmark but fail to leverage the universal statistics in the sign language domain.

Notably, self-supervised pre-training techniques represented by BERT~\cite{devlin2018bert} have achieved great success in Natural Language Processing~(NLP).
BERT builds on the strong Transformer~\cite{vaswani2017attention} backbone and designs an ingenious pretext task, \emph{i.e.,} masked language modeling~(MLM).
It aims to reconstruct the masked word tokens from the corrupted input sequence, whose objective is implemented by cross entropy to maximize the joint word probability distribution.
However, the main obstacle to leverage its success in video SLR is the different characteristics of the input signal.
In NLP, the input word token is discrete and pre-defined with high semantics.
In contrast, the video signal of sign language is continuous with the spatial and temporal dimensions.
This signal is quite low-level, making the original BERT objective not applicable.
Besides, since the sign language video is mainly characterized by hand and body movements, the direct adoption of the BERT framework may not be optimal.

To tackle the above issue, we propose a self-supervised pre-trainable framework with a specific design for sign language, namely BEST.
Focusing on the main properties of sign language, we organize the hand and body as the pose triplet unit.
This triplet unit is embedded and fed into the Transformer backbone.
Basically, our framework contains two stages, \emph{i.e.,} self-supervised pre-training and downstream fine-tuning.
During pre-training, we propose the masked unit modeling~(MUM) pretext task to capture the context cues.
The input hand or body unit embedding is randomly masked, and then the framework reconstructs the masked unit from this corrupted input sequence.
Similar to BERT, self-reconstruction is optimized via the cross-entropy objective.
To this end, we jointly tokenize the pose triplet unit as the pseudo label, which represents the gesture/body state.
After pre-training, the pre-trained Transformer encoder is fine-tuned with the newly added prediction head to perform the SLR task.

Our contributions are summarized as follows,
\begin{itemize}
    \item We propose a self-supervised pre-trainable framework.
    It leverages the BERT success, jointly with the specific design for the sign language domain. 
    \item We organize the main hand and body movement as the pose triplet unit and propose the masked unit modeling~(MUM) pretext task. To utilize the BERT objective, we generate the pseudo label for this task via coupling tokenization on the pose triplet unit. 
    \item Extensive experiments on downstream SLR validate the effectiveness of our proposed method, achieving new state-of-the-art performance on four benchmarks with a notable gain.
\end{itemize}

\section{Related Work}
In this section, we will briefly review several related topics, including sign language recognition and self-supervised pre-training.

\subsection{Sign Language Recognition}
Sign language recognition has received much attention in recent years~\cite{koller2018deep,momeni2022automatic,niu2020stochastic,jin2021contrastive,hu2021hand,li2020transferring}.
Typically, the research works can be grouped into two categories based on the input modality, \emph{i.e.,} RGB-based methods and pose-based methods.

\textbf{RGB-based Methods.} 
Early works~\cite{farhadi2007transfer,fillbrandt2003extraction,starner1995visual} on SLR focused on hand-crafted features computed for hand shape variation and body motion. 
Along with the popularity of convolutional neural networks~(CNNs) in computer vision, many works in SLR adopt CNNs as the backbone~\cite{selvaraj-etal-2022-openhands,sincan2020autsl,joze2018ms,koller2018deep,hu2021hand}. 
For example, 3D CNNs are adopted due to their representation capacity for spatio-temporal dependency~\cite{huang2018attention,joze2018ms,albanie2020bsl,li2020transferring}.

\textbf{Pose-based Methods.} 
Pose modality is a compact and high-level representation of human action and contains physical connection among skeleton joints~\cite{li2018co,ng2021body2hands,yan2018spatial}. 
To extract the semantic representation of pose data, some works~\cite{li2018co,du2015hierarchical} explore graph convolutional networks~(GCNs) as the backbone. 
These GCN-based methods~\cite{camgoz2018neural,li2018co,yan2018spatial,min2020efficient} show impressive performance in action recognition. 
Tunga \textit{et al.}~\cite{tunga2021pose} combine GCN and Transformer to capture spatial-temporal information based on sign language pose sequence for sign language recognition. 
In this work, given the compactness of pose data, we utilize them as our input modality.

\subsection{Self-supervised Pre-Training}
Self-supervised pre-training methods have achieved remarkable success in Natural Language Processing~(NLP) and Computer Vision~(CV) fields, which make full use of large-scale unlabeled data to learn generic feature representation for a wide range of downstream tasks.
In NLP, with the strong modeling capability of Transformer~\cite{vaswani2017attention}, many works propose to pre-train on this backbone for generic representations~\cite{devlin2018bert,lewis2019bart,conneau2019cross}. 
BERT is one of the most popular methods, which designs a cleverly masked language modeling~(MLM) pretext task. 
MLM predicts the masked word tokens from the corrupted input sentence, which aims to capture the context cues in the text corpus.

Motivated by BERT, some works attempt to leverage its success into CV tasks~\cite{su2019vl,sun2019videobert,zhu2020actbert,bao2021beit,he2022masked}.
There exist different characteristics of the input signal between NLP and CV.
Different from the semantic discrete word token, the signal in CV tasks is usually low-level and continuous, which makes the original BERT objective not applicable.
One way to tackle this problem is changing its objective into regression.
He \textit{et al.}~\cite{he2022masked} propose masked autoencoders~(MAE) to reconstruct the missing pixels of masked image patches using regression objective.
Hu \textit{et al.}~\cite{hu2021signbert} propose a pre-trained model for sign language based on self-reconstruction of the hand pose data.
Jiang \textit{et al.}~\cite{jiang2021skeletor} attempt to utilize BERT-style refined pose for sign language. 
According to~\cite{ramesh2021zero}, the former method may focus too many short-range dependencies and hurt the downstream performance.
Therefore, some methods turn to tokenizing the input signal to provide the discrete pseudo label.
BEiT~\cite{bao2021beit} propose masked image modeling~(MIM) with tokenized image patches as supervision.
It is originally designed for image-based tasks, and cannot be directly adopted into the video-based sign language domain, due to different task characteristics and input modalities.

\section{Methods}

\begin{figure*}[t]
	\centering
	\includegraphics[width=0.9\linewidth]{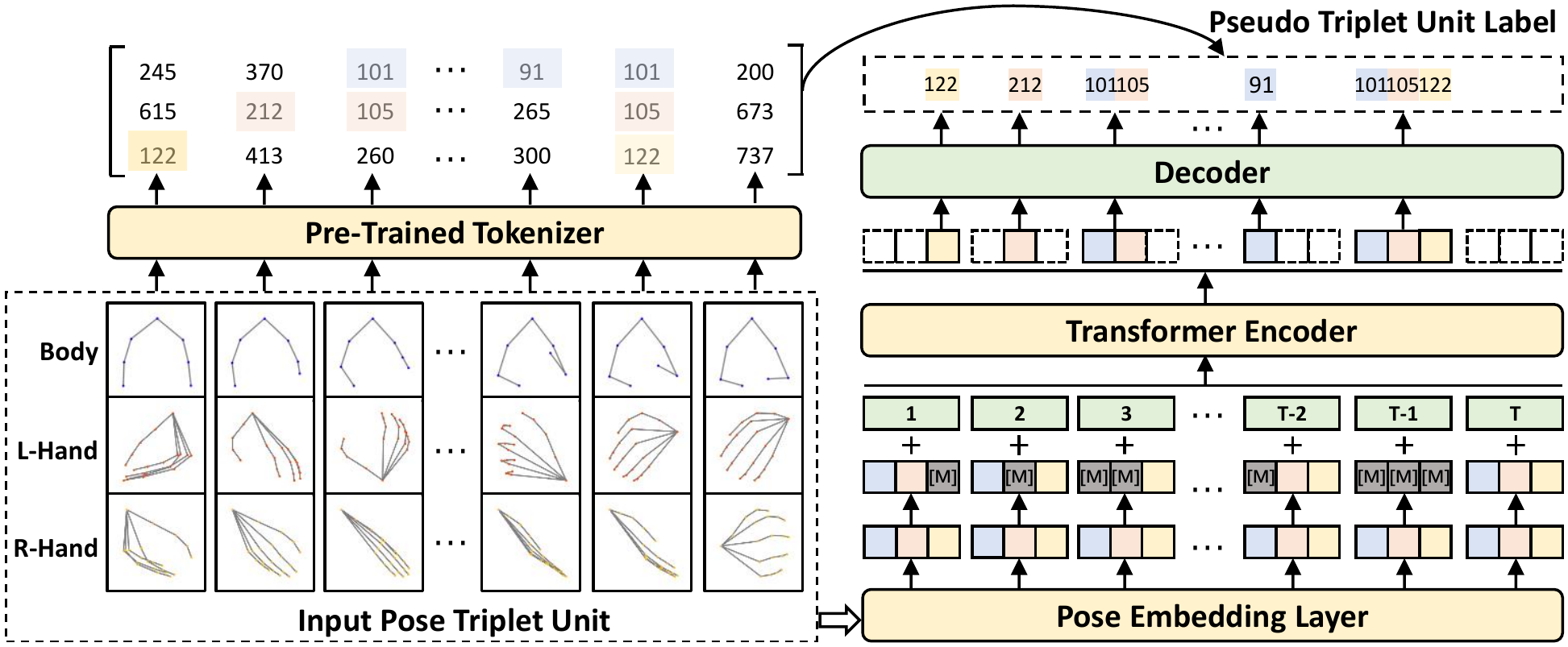}
	\caption{Illustration of our proposed BEST framework during pre-training. BEST mainly contains four components, \textit{i.e.,} a pre-trained tokenizer, a pose embedding layer, a Transformer encoder, and a decoder. The input pose triplet unit is composed by the body pose, the left and right hand pose. Given a sequence of pose triplet units, we utilize the proposed MUM pretext task to pre-train our framework. The [M] denotes the learnable masked token for the input sequence. Meanwhile, the pre-trained tokenizer coupling discretizes the triplet units into pseudo triplet unit labels to supervise the pre-training procedure.}
	\label{fig:overview}
	\vspace{-1.5em}
\end{figure*}

Our proposed method consists of two stages, \textit{i.e.,} self-supervised pre-training and downstream task fine-tuning.
As shown in Figure~\ref{fig:overview}, during pre-training, we first utilize a discrete variational autoencoder to learn codebooks for the upper body, left and right hand.
It performs coupling tokenization on the pose triplet unit as the pseudo label for the following pretext task.
Then we pre-train the BEST model via our designed masked unit modeling~(MUM) pretext task to capture the context cues.
Finally, we append the task-specific layer and fine-tune the pre-trained parameters on the downstream SLR task.

\subsection{Tokenization in Pre-Training}\label{Tokenization}
The tokenization provides pseudo labels for our designed pretext task during pre-training.
We utilize a discrete variational autoencoder~(d-VAE) to jointly convert the pose triplet unit into the triplet tokens~(body, left and right hand), motivated by VQ-VAE~\cite{van2017neural}.
Our utilized pose triplet unit $\mathnormal{J}_\mathit{sign}$ consists of two hand poses $\mathnormal{J}_\mathit{left},  \mathnormal{J}_\mathit{right}$, and an upper body pose $\mathnormal{J}_\mathit{body}$. 
The d-VAE contains three parts, \emph{i.e.,} encoder, quantizer, and decoder.
The encoder maps the pose triplet unit to the intermediate latent vector $\mathit{z} = \mathrm{Enc}(\mathnormal{J}_\mathit{sign})$. 
The quantizer is in charge of tokenizing each vector to be codewords coming from the hand codebook $\mathcal{V}_{hand} = \{\mathit{h}_{k}\}_{k=1}^{M_1}$ and the upper body codebook $\mathcal{V}_{body} = \{\mathit{d}_{k}\}_{k=1}^{M_2}$. 
The quantized vector $\mathit{z}_\mathbf{q}$ is computed as follows, 

\begin{equation}
	\label{equ1}
	\begin{split}
		\mathit{z}_\mathbf{q} &= \mathrm{Concat}(\mathit{h}_{k^l}, \mathit{h}_{k^r}, \mathit{d}_{k^b}),
		\mathit{z} =  \mathrm{Concat}(\mathit{z}_{l}, \mathit{z}_{r}, \mathit{z}_{b}),  \\
		k^l &= \mathrm{Q\_hand}(\mathit{z}_l) = \mathop{\arg\min}\limits_{k \in \{1,2, \cdots, M_1\}} ||\mathit{z}_l - \mathit{h}_{k}||, \\
		k^r &= \mathrm{Q\_hand}(\mathit{z}_r) = \mathop{\arg\min}\limits_{k \in \{1,2, \cdots, M_1\}} ||\mathit{z}_r - \mathit{h}_{k}||, \\
		k^b &= \mathrm{Q\_body}(\mathit{z}_b) = \mathop{\arg\min}\limits_{k \in \{1,2, \cdots, M_2\}} ||\mathit{z}_b - \mathit{d}_{k}||,
	\end{split}
\end{equation}	
where $l, r$ and $b$ indicate the abbreviation of the `left', `right' and `upper body', respectively. 
We still adopt the abbreviations $l, r$ and $b$ in the following introduction.
The latent vector $\mathit{z}$ is made up of three local representations of the pose triplet unit, \textit{i.e.,} $\mathit{z}_l, \mathit{z}_r$, and $\mathit{z}_b$.
$\mathrm{Q\_hand}(\cdot)$ is the hand quantization encoder that maps the vector to the index of the hand codebook, which satisfies the minimum semantic distance between the input vector and the corresponding codeword.  
Similar to the hand quantizer, $\mathrm{Q\_body}(\cdot)$ denotes the body quantization encoder for the upper body. 
The quantized vector $\mathit{z}_\mathbf{q}$ consists of a cascade of three codewords from two codebooks. Based on the quantized vector, the decoder aims to reconstruct the input pose triplet unit. The output of the decoder is computed as follows,
 
\begin{equation}
	\label{equ2}
	\hat{\mathnormal{J}}_\mathit{sign} = \mathrm{Dec}(\mathit{z}_\mathbf{q}),
\end{equation}

\noindent whose output is the reconstructed hand and body poses~($\hat{\mathnormal{J}}_\mathit{left}, \hat{\mathnormal{J}}_\mathit{right}$ and $\hat{\mathnormal{J}}_\mathit{body}$). 

This tokenizer is trained before self-supervised pre-training.
However, the tokenization process is non-differentiable.
To backpropagate the gradient from the decoder to the encoder, we utilize the straight-through estimator~\cite{bengio2013estimating} to copy the gradient from the decoder to the encoder. 
The training objective function of d-VAE is defined as follows,

\begin{equation}
	\label{equ3}
	\begin{aligned}
	\mathcal{L}_{d-VAE} &= \mathcal{L}_{hand} + \beta_{1} \mathcal{L}_{body} + \beta_{2} ||\mathbf{sg}[\mathit{z}] - \mathit{z}_\mathbf{q} ||_2^2  \\ &+ \beta_{3} ||\mathbf{sg}[\mathit{z}_\mathbf{q}] - \mathit{z} ||_2^2, \\
	\mathcal{L}_{hand} &= || \hat{\mathnormal{J}}_\mathit{left} - \mathnormal{J}_\mathit{left} || + || \hat{\mathnormal{J}}_\mathit{right} - \mathnormal{J}_\mathit{right} ||, \\
	\mathcal{L}_{body} &= || \hat{\mathnormal{J}}_\mathit{body} - \mathnormal{J}_\mathit{body} ||,
	\end{aligned}
\end{equation}
where $\mathcal{L}_{hand}$ and  $\mathcal{L}_{body}$ denote the reconstruction losses of 2D hand and upper body pose, respectively. 
$\mathbf{sg}[\cdot]$ represents the stop-gradient operator, and $\beta_{1}, \beta_{2}$ and $\beta_{3}$ are the weighting factors to balance the impact of the four losses.

\subsection{Framework Architecture in Pre-Training}
During pre-training, the framework contains four components, \emph{i.e.,} a pose embedding layer, a pre-trained pose tokenizer, a Transformer encoder, and a decoder.

\textbf{Pose Embedding Layer.} 
This embedding layer aims to extract the feature embedding from the input pose triplet unit.
Since pose has physical connection relationship, we utilize the graph convolutional network~(GCN) proposed in~\cite{cai2019exploiting} as the pose embedding layer. 
Specifically, given a pose sequence $\mathnormal{V}_{\mathit{sign}} = \{\mathnormal{J}_{\mathit{sign}, t}\}_{t=1}^T$ with $T$ frames, it extracts the pose triplet unit $\mathit{f}_{\mathit{sign}, t} \in \mathbb{R}^\mathit{D}$ from the body, left and right hand in a frame-wise manner and concatenate them for the following module. 
Each part of the triplet unit has the same feature dimension $D_{\textit{part}} = \frac{1}{3}D$. 

\textbf{Tokenizer.}  
As presented in Section~\ref{Tokenization}, given a pose sequence $ \{\mathnormal{J}_{\mathit{sign}, t}\}_{t=1}^T$, we tokenize it as the pseudo labels $\{\mathit{k}_t\}_{t=1}^T$ for pre-training, of which $\mathit{k}_t$ denotes the concatenation of $\mathit{k}^l_{t}$, $\mathit{k}^r_{t}$ and $\mathit{k}_t^b$ computed following Equation~\eqref{equ1}. 
The tokenizer is only adopted for pseudo label inference, with all parameters frozen during pre-training.

\textbf{Transformer Encoder.} 
Given the feature sequence from the pose embedding layer $\mathrm{F}_{\mathit{sign}} = \{\mathit{f}_{\mathit{sign}, t}\}_{t=1}^T$, 
we further add temporal embedding $\mathit{f}_{\mathit{temp}, t} \in \mathbb{R}^D$ implemented by the position encoding~\cite{vaswani2017attention}. 
The input sequence $\mathrm{F}_{\footnotesize{0}}$ is computed as follows,

\begin{equation}
	\label{equ4}
	\begin{split}
		\mathrm{F}_{\footnotesize{0}} &= [\mathnormal{f}_{in, 1}, \cdots , \mathnormal{f}_{in, T}] , \\
		\mathnormal{f}_{in, t} &= \mathnormal{f}_{mask, t} + \mathnormal{f}_{temp, t} , \quad t \in \{1, \cdots, T\} , \\
		\mathrm{F}_{\mathit{m}} &= \large{\mathrm{Mask}}(\mathrm{F}_{\mathit{sign}}) = [\mathnormal{f}_{mask, 1}, \cdots , \mathnormal{f}_{mask, T}] ,
	\end{split}
\end{equation}

\noindent where $\large{\mathrm{Mask}}(\cdot)$ denotes the masking operator for the embedding sequence and the masked frame positions are denoted as $\mathcal{M} \in \{1,\cdots, T\}^{\alpha \cdot T}$, in which $\alpha$ denotes the mask ratio.  
We will explain the masked modeling in Section~\ref{pretext}. 
The input sequence $\mathrm{F}_{\footnotesize{0}}$ is fed into the Transformer encoder. 
The Transformer encoder contains $N$ layers of Transformer blocks $ \mathrm{F}_l = Block(\mathrm{F}_{l-1})$, where $l = 1, \cdots, N$.
The output sequence of the last layer $\mathrm{F}_N = [\mathnormal{f}_{out, 1}, \cdots, \mathnormal{f}_{out, T}]$ is utilized as the encoded representation of the pose triplet unit, where $\mathnormal{f}_{out, t}$ corresponds to the $t$-th frame.

\textbf{Decoder.} 
Given the output sequence $\mathrm{F}_N$, the  masked frame positions $\mathcal{M}$ and the pseudo labels $\{\mathit{k}_t\}_{t=1}^T$, we utilize the decoder to reconstruct the pose triplet unit.
Since each output feature $\mathit{f}_{out,t}$ contains three parts, \textit{i.e.,} $\mathit{f}_{out,t}^l$, $\mathit{f}_{out,t}^r$ and $\mathit{f}_{out,t}^b$, we record the masked positions of each part into $\mathcal{M}^l$, $\mathcal{M}^r$ and $\mathcal{M}^b$, respectively. $\mathcal{M} = \mathcal{M}^l \cup \mathcal{M}^r \cup \mathcal{M}^b$. For each feature $\mathit{f}_{\textit{out}, t}$, we use the softmax classifier to predict the corresponding label of each part,
\begin{equation}
	\label{equ5}
	\begin{split}
		p_{\mathrm{hand}}(k_t^{l} | \mathit{f}_{out,t}) &= \mathrm{softmax}_{k_t^{l}}(\mathnormal{W}_{1}\mathit{f}_{out,t}^l + \mathnormal{b}_1), \quad t \in \mathcal{M}^l, \\
		p_{\mathrm{hand}}(k_t^{r} | \mathit{f}_{out,t}) &= \mathrm{softmax}_{k_t^{r}}(\mathnormal{W}_{1}\mathit{f}_{out,t}^r + \mathnormal{b}_1), \quad t \in \mathcal{M}^r, \\
		p_{\mathrm{body}}(k_t^{b} | \mathit{f}_{out,t}) &= \mathrm{softmax}_{k_t^{b}}(\mathnormal{W}_{2}\mathit{f}_{out,t}^b + \mathnormal{b}_2), \quad t \in \mathcal{M}^b,\\
	\end{split}
\end{equation}

\noindent where $l$, $r$ and $b$ denote the abbreviation of left, right and upper body, respectively. $\mathnormal{W}_{1} \in \mathbb{R}^{|\mathcal{V}_{hand}| \times \mathit{D}_{\textit{part}}}$ and $\mathnormal{W}_{2} \in \mathbb{R}^{|\mathcal{V}_{body}| \times \mathit{D}_{\textit{part}}}$ denote the projection matrix.  $\mathnormal{b}_{1} \in \mathbb{R}^{|\mathcal{V}_{hand}|}$ and $\mathnormal{b}_{2} \in \mathbb{R}^{|\mathcal{V}_{body}|}$ denote the biases.

\subsection{Pretext Task in Pre-Training \& Objective}
\label{pretext}
Our designed pretext task is MUM, which aims to exploit the hierarchical correlation context among internal and external triplet pose units.
Given a pose sequence with a triplet pose unit of length $T$, we first randomly choose the $\alpha \cdot T$ frames to process the mask operation.
For clarification, we define three parts of the pose triplet unit as $\mathit{f}_{\mathit{sign}, t}^l$, $\mathit{f}_{\mathit{sign}, t}^r$ and $\mathit{f}_{\mathit{sign}, t}^b$, respectively. 
If a unit is masked, a learnable masked token $\mathbf{e}_{mask} \in \mathbb{R}^{\mathit{D}_\mathit{part}}$ is utilized to replace each part of the triplet unit with 50\% probability. 
Therefore, the masked triplet unit includes three masking cases: only hand masked, only body masked and hand-body masked.
The hand masked case aims to endow our model with the capacity of capturing the local hand context. 
Compared with the hand counterpart, the body masked case aims to mine the context of global body movements. 
The hand-body masked case means that body and hand features are masked simultaneously. 
Since sign language conveys the full meaning with the cooperation of hand and body, we expect to exploit the correlation context cues between hand and body in this case.
Therefore, we utilize the MUM to pre-train our method and conduct an ablation experiment to validate our thoughts.

The overall pre-training objective is to maximize the log-likelihood of the correct labels given the corrupted sequence, which is computed as follows,
\begin{equation}
	\label{equ6}
	\begin{split}
	\mathcal{L}_{\mathrm{pre-train}} &= \mathrm{max} \sum\limits_{\mathnormal{V}_{\mathit{sign}} \in \mathcal{D}} \mathbb{E}_{\mathcal{M}} \left[\sum\limits_{t \in \mathcal{M}} \log p(k_t|\mathnormal{V}_{\mathit{sign}})\right] \\
	\sum\limits_{t \in \mathcal{M}} \log p(k_t|\mathnormal{V}_{\mathit{sign}}) &= \sum\limits_{t \in \mathcal{M}^l} \log p_{\mathrm{hand}}(k_t^l|\mathit{f}_{\mathit{out}, t}^l) \\ &+ \sum\limits_{t \in \mathcal{M}^r} \log p_{\mathrm{hand}}(k_t^r|\mathit{f}_{\mathit{out}, t}^r) \\
	&+ \sum\limits_{t \in \mathcal{M}^b} \log p_{\mathrm{body}}(k_t^b|\mathit{f}_{\mathit{out}, t}^b) ,
\end{split}
\end{equation}
\noindent where $\mathcal{D}$ is the training corpus, $\mathcal{M}$ represents the masked frame positions, $\mathit{f}_{out,t}$ is the masked triplet unit. $\mathcal{M}^l$ and $\mathcal{M}^r$ and $\mathcal{M}^b$ represent the masked positions for left hand, right hand and body, respectively.

\subsection{Downstream Fine-Tuning}
After pre-training, we directly fine-tune the parameters under the downstream SLR task. 
We replace the decoder with a simple MLP as the prediction head. 
During fine-tuning, we do not mask any triplet pose unit, and supervise the output of the prediction head with ground-truth labels. 
Besides, only the pose information is insufficient to convey the full meaning of sign language. 
We utilize a late fusion strategy to sum the prediction results of pose data and RGB data. 
In our experiment, we refer our method with only pose data, fusion of RGB data as \textbf{Ours} and \textbf{Ours~(+R)}, respectively.

\section{Experiments}
\subsection{Datasets and Metrics} \label{dataset_intro}

\textbf{Datasets.} 
We conduct experiments on four public sign language datasets, \textit{i.e.,} NMFs-CSL~\cite{hu2021global}, SLR500~\cite{huang2018attention}, WLASL~\cite{li2020word} and MSASL~\cite{joze2018ms}. The training sets of all datasets participate in the pre-training stage.
Table~\ref{datasets} presents an overview of the above-mentioned datasets.

NMFs-CSL is a large-scale Chinese sign language~(CSL) dataset with a vocabulary size of 1,067. 
All samples are split into 25,608 and 6,402 samples for training and testing, respectively. 
SLR500 is another CSL dataset including 500 daily words performed by 50 signers. 
It contains a total of 125,000 samples, of which 90,000 and 35,000 samples are utilized for training and testing, respectively. 
The above two datasets collect samples in the controlled lab scene.

\begin{table}[t]
	\footnotesize
	\tabcolsep=3.5pt
	\begin{center}
	    \resizebox{\linewidth}{!}{
		\begin{tabular}{llllll}
			\toprule
			\textbf{Name} & \textbf{Language} & \textbf{Vocab.} & \textbf{Videos} & \textbf{Signers} & \textbf{Source}  \\ \midrule
			WLASL~\cite{li2020word} & ASL & 2000 & 21.1k & 119 &  Web, lexicons  \\
			NMF-CSL~\cite{hu2021global} & CSL & 1067 & 32.0k & 10 &  lab \\
			MSASL~\cite{joze2018ms} & ASL & 1000 & 25.5k & 222 & Web, lexicons \\  
			SLR500~\cite{huang2018attention} & CSL & 500 & 125k & 50 & lab \\ \bottomrule
		\end{tabular}
		}
	\end{center}
	\vspace{-1.0em}
         \caption{Statistics of public isolated SLR datasets. ASL denotes American Sign Language, and CSL denotes Chinese Sign Language.}
	\label{datasets}
        \vspace{-1.5em}
\end{table}

WLASL is a large-scale American sign language~(ASL) dataset, containing 2000 words performed by over 100 signers. 
It totally consists of 21,083 samples. 
In particular, it selects the top-$K$ most frequent words with $K = \{100, 300\}$, and organize them as two subsets, namely WLASL100 and WLASL300. 
MSASL is another popular ASL dataset with a total of 25,513 samples and a vocabulary size of 1000. 
Similar to WLASL, it also provides two subsets, named MSASL100 and MSASL200, respectively. 
Different from NMFs-CSL and SLR500, WLASL and MSASL collect data from the Web and bring more challenges due to the unrestricted real-life scenario. \\

\noindent \textbf{Metrics.} 
For evaluation, we report the classification accuracy, \textit{i.e.,} Top-1~(\textbf{T-1}) and Top-5~(\textbf{T-5}) for the downstream SLR task. 
We adopt both per-instance~(\textbf{P-I}) and per-class~(\textbf{P-C}) accuracy metrics following~\cite{li2020word,joze2018ms}. 
Per-instance accuracy is computed over whole test data. 
Per-class accuracy is the average of the sign categories present in the test set. 
For WLASL and MSASL, we report both per-instance and per-class accuracy due to the unbalanced samples per class of the two datasets. 
For NMFs-CSL and SLR500, we only report per-instance accuracy with an equal number of samples for each class.

\subsection{Experiment Settings}

\textbf{Data Preparation and Processing.} 
Our proposed method utilizes the pose modality to represent hand and body movements. 
Since no available pose annotation is provided in sign language datasets, we utilize the off-the-shelf pose detector MMPose~\cite{mmpose2020} to extract the 2D pose keypoints. 
In each frame, the extracted 2D pose includes 49 joints, containing 7 upper body joints and 42 hand joints. 
Furthermore, considering the different scales among triplet units, we crop the body, left and right hand pose separately and rescale them to 256 $\times$ 256.

\noindent \textbf{Model Hyper-Parameters.} 
For the tokenizer, the vocabulary size of hand codebook $M_1$ and body codebook $M_2$ are 1000 and 500, respectively.  
The dimension of each codeword in two codebooks is 512. 
The weighting factors $\beta_{1}$, $\beta_{2}$ and $\beta_{3}$ in equation~\ref{equ3} are set to 0.1, 1.0 and 0.9, respectively.
During pre-training, the Transformer encoder contains 8 heads with the input size of the Transformer encoder $D$ as 1536 and position-wise feed-forward dimension as 2048.

\noindent \textbf{Training Setup.} 
The Adam~\cite{kingma2014adam} optimizer is employed in our experiments.
For tokenizer training, we set the initial learning rate as 0.001 and decrease it with a factor of 0.1 per 10 epochs. For pre-training, the weight decay and momentum are set to 0.01 and 0.9, respectively. The learning rate is set to 0.0001, with a warmup of 6 epochs, and linear learning rate decay. 
For the downstream SLR task, the learning rate is initialized to 0.0001 and decreases by a factor of 0.1 per 10 epochs. we disturb the coordinates of the input pose sequence with a random perturbation matrix to relieve overfitting during training. 
Following~\cite{hu2021signbert}, we temporally select 32 frames using random and center temporal sampling during training and testing, respectively. 
All experiments are implemented by PyTorch and trained on NVIDIA RTX 3090.

\begin{table*}[t!]
	\footnotesize
	\tabcolsep=10.0pt
	\begin{center}
		\resizebox{\textwidth}{!}{
		\begin{tabular}{lcccc|cccc|cccc}
			\toprule
			\multirow{3}{*}[-0.2in]{Method} & \multicolumn{4}{c|}{MSASL100}
			& \multicolumn{4}{c|}{MSASL200}
			& \multicolumn{4}{c}{MSASL1000}\\ 
			& \multicolumn{2}{c}{P-I} & \multicolumn{2}{c|}{P-C} 
			& \multicolumn{2}{c}{P-I} & \multicolumn{2}{c|}{P-C}
			& \multicolumn{2}{c}{P-I} & \multicolumn{2}{c}{P-C} \\ 
			\cmidrule(lr){2-3} \cmidrule(lr){4-5} \cmidrule(lr){6-7} \cmidrule(lr){8-9} \cmidrule(lr){10-11} \cmidrule(lr){12-13}
			& T-1 & T-5 & T-1 & \multicolumn{1}{c|}{T-5}  
			& T-1 & T-5 & T-1 & \multicolumn{1}{c|}{T-5} 
			& T-1 & T-5 & T-1 & \multicolumn{1}{c}{T-5}    \\ \toprule
			\textbf{Pose-based} & & & & & & & & & & & & \\
			ST-GCN~\cite{yan2018spatial} & 59.84 & 82.03 & 60.79 & 82.96  
			& 52.91 & 76.67 & 54.20 & 77.62
			& 36.03 & 59.92 & 32.32 & 57.15 \\ 
			SignBERT~\cite{hu2021signbert} & 76.09 & 92.87 & 76.65 & 93.06  
			& 70.64 & 89.55 & 70.92 & 90.00
			& 49.54 & 74.11 & 46.39 & 72.65 \\
			Ours & \textbf{80.98} & \textbf{95.11} & \textbf{81.24} & \textbf{95.44} & \textbf{76.60} &\textbf{91.54} &\textbf{76.75} &\textbf{91.95} & \textbf{58.82} &\textbf{81.18} & \textbf{54.87} & \textbf{80.05}\\  
   \midrule
			\textbf{RGB-based} & & & & & & & & & & & &   \\
			I3D~\cite{carreira2017quo}  & - & - & 81.76 & 95.16  
			& - & - & 81.97 & 93.79
			& - & - & 57.69 & 81.05\\ 
			TCK~\cite{li2020transferring}  & 83.04 & 93.46 & 83.91 & 93.52  
			& 80.31 & 91.82 & 81.14 & 92.24
			& - & - & - & - \\ 
			BSL~\cite{albanie2020bsl}  & - & - & - & -  
			& - & - & - & -
			& 64.71 & 85.59 & 61.55 & 84.43 \\
			HMA~\cite{hu2021hand} & 73.45 & 89.70 & 74.59  & 89.70 &66.30  & 84.03 & 67.47 & 84.03 & 49.16 & 69.75  & 46.27 & 68.60 \\
			Ours~(+R) & \textbf{89.56} & \textbf{96.96} & \textbf{90.08} & \textbf{97.07} & \textbf{86.83} & \textbf{95.66} & \textbf{87.45} & \textbf{95.72} & \textbf{71.21} & \textbf{88.85} & \textbf{68.24} & \textbf{87.98} \\ \bottomrule
		\end{tabular}
	}
	\end{center}
        \vspace{-1em}
         \caption{Comparison with state-of-the-art methods on MSASL dataset. Our proposed method fused with another RGB-based method, I3D~\cite{carreira2017quo}, is represented by Ours~(+R).}
        \vspace{-1em}
	\label{msasl}
\end{table*}
\begin{table*}[t!]
	\footnotesize
	\tabcolsep=9.0pt
	\begin{center}
		\resizebox{\textwidth}{!}{
		\begin{tabular}{lcccc|cccc|cccc}
			\toprule
			\multirow{3}{*}[-0.2in]{Method} & \multicolumn{4}{c|}{WLASL100}
			& \multicolumn{4}{c|}{WLASL300}
			& \multicolumn{4}{c}{WLASL2000}\\ 
			& \multicolumn{2}{c}{P-I} & \multicolumn{2}{c|}{P-C} 
			& \multicolumn{2}{c}{P-I} & \multicolumn{2}{c|}{P-C}
			& \multicolumn{2}{c}{P-I} & \multicolumn{2}{c}{P-C} \\ 
			\cmidrule(lr){2-3} \cmidrule(lr){4-5} \cmidrule(lr){6-7} \cmidrule(lr){8-9} \cmidrule(lr){10-11} \cmidrule(lr){12-13}
			& T-1 & T-5 & T-1 & \multicolumn{1}{c|}{T-5}  
			& T-1 & T-5 & T-1 & \multicolumn{1}{c|}{T-5} 
			& T-1 & T-5 & T-1 & \multicolumn{1}{c}{T-5}    \\ \toprule
			\textbf{Pose-based} & & & & & & & & & & & & \\
			ST-GCN~\cite{yan2018spatial}& 50.78 & 79.07 & 51.62 & 79.47   
			& 44.46 & 73.05 & 45.29 & 73.16
			& 34.40 & 66.57 & 32.53 & 65.45 \\ 
			Pose-TGCN~\cite{li2020word} & 55.43 & 78.68 & - & -   
			& 38.32 & 67.51 & - & -
			& 23.65 & 51.75 & - & -\\ 
			PSLR~\cite{tunga2021pose}& 60.15 & 83.98 & - & -   
			& 42.18 & 71.71 & - & -
			& - & - & - & - \\
			SignBERT~\cite{hu2021signbert} & 76.36 & 91.09 & 77.68 & 91.67  
			& 62.72 & 85.18 & 63.43 & 85.71 
			& 39.40 & 73.35 & 36.74 & 72.38  \\ 
			Ours & \textbf{77.91} & \textbf{91.47} & \textbf{77.83} & \textbf{92.50} & \textbf{67.66} & \textbf{89.22} & \textbf{68.31} & \textbf{89.57} & \textbf{46.25} & \textbf{79.33} & \textbf{43.52} & \textbf{77.65} \\ 
\midrule
			\textbf{RGB-based} & & & & & & & & & & & & \\
			I3D~\cite{carreira2017quo}& 65.89 & 84.11 & 67.01 & 84.58  
			& 56.14 & 79.94 & 56.24 & 78.38
			& 32.48 & 57.31 & - & -\\ 
			TCK~\cite{li2020transferring} & 77.52 & 91.08 & 77.55 & 91.42   
			& 68.56 & 89.52 & 68.75 & 89.41
			& - & - & - & - \\ 
			BSL~\cite{albanie2020bsl} & - & - & - & -  
			& - & - & - & -
			& 46.82 & 79.36 & 44.72 & 78.47\\
			HMA~\cite{hu2021hand} & - & - & - & -
			& - & - & - & - 
			& 37.91 & 71.26 & 35.90 & 70.00 \\
			Ours~(+R) & \textbf{81.01} & \textbf{94.19} & \textbf{81.63} & \textbf{94.67} & \textbf{75.60} & \textbf{92.81} & \textbf{76.12} & \textbf{93.07} & \textbf{54.59} & \textbf{88.08} & \textbf{52.12} & \textbf{87.28} \\ \bottomrule 
		\end{tabular}
	}
	\end{center}
        \vspace{-1em}
        \caption{Comparison with state-of-the-art methods on WLASL dataset. Our proposed method fused with another RGB-based method, I3D~\cite{carreira2017quo}, is represented by Ours~(+R).}
	\label{wlasl}
	\vspace{-1.5em}
\end{table*}

\subsection{Comparison with State-of-the-art Methods}
In this section, we compare our method with previous state-of-the-art methods on four public datasets. 
Following~\cite{hu2021signbert}, we group the previous methods by the input modality, \textit{i.e.,} pose-based and RGB-based methods.

\noindent \textbf{NMFs-CSL Dataset.} 
As shown in Table~\ref{NMFs-CSL}, we compare with the previous methods.
SignBERT~\cite{hu2021signbert} is a self-supervised pre-training method with hand prior to model the hand sequence. 
Compared with SignBERT, our method still achieves the best performance in top-1 accuracy under different settings. 
GLE-Net~\cite{hu2021global} is the state-of-the-art method with discriminative clues from global and local views. 
Compared with GLE-NET, Ours~(+R) outperforms it, achieving 79.2\% top-1 accuracy.

\noindent \textbf{MSASL Dataset.} 
As shown in Table~\ref{msasl}, ST-GCN~\cite{yan2018spatial} shows inferior performance compared with RGB-based methods, which may be attributed to the failure of pose detection. 
Compared with ST-GCN~\cite{yan2018spatial}, self-supervised learning methods, \textit{i.e.,} SignBERT~\cite{hu2021signbert} and Ours, relieve this issue by leveraging the stored statistics during pre-training. 
Our method outperforms SignBERT~\cite{hu2021signbert} with 4.89\%, 5.96\% and 9.28\% Top-1 per-instance accuracy improvement on MSASL100, MSASL200 and MSASL1000, respectively.
Notably, our method even achieves comparable performance with RGB-based methods.
When fused with the RGB method, the performance is further improved.

\noindent \textbf{WLASL Dataset.} 
Similar to MSASL, WLASL is a challenging dataset with unconstrained recording conditions and unbalanced sample distribution. 
As shown in Table~\ref{wlasl}, Pose-TGCN~\cite{li2020word} and PSLR~\cite{tunga2021pose} show inferior performance caused by the erroneous estimation of poses. 
HMA~\cite{hu2021hand} utilizes a hand statistic model to refine the pose and improve the performance. 
BSL~\cite{albanie2020bsl} and TCK~\cite{li2020transferring} utilize external RGB data to enhance the model robustness and boost their performance. 
Compared with them, our method achieves the best performance on all subsets.

\noindent \textbf{SLR500 Dataset.} 
As shown in Table~\ref{slr500}, compared with STIP~\cite{laptev2005space} and GMM-HMM~\cite{tang2015real} based on hand-craft features, deep-learning-based methods~\cite{yan2018spatial,qiu2017learning,hu2021signbert} achieve notable performance gain.
Our method still achieves new state-of-the-art performance among both pose-based and RGB-based methods.

\begin{table}[t]
	\centering
	\tabcolsep=35pt
	\resizebox{0.99\linewidth}{!}{
	\begin{tabular}{lc}
		\toprule
		Method  &  Accuracy   \\  \toprule
		\textbf{Pose-based} \\
		ST-GCN~\cite{yan2018spatial} &  90.0 \\ 
		SignBERT~\cite{hu2021signbert}      & 94.5     \\ 
		Ours  &  \textbf{95.4}   \\ 
        \midrule
		\textbf{RGB-based}  \\
		STIP~\cite{laptev2005space}   &  61.8 \\
		GMM-HMM~\cite{tang2015real} &  56.3 \\
		3D-R50~\cite{qiu2017learning} &  95.1 \\
		HMA~\cite{hu2021hand} & 95.9 \\
		GLE-Net~\cite{hu2021global}   & 96.8      \\
		Ours~(+R) & \textbf{97.7} \\ \bottomrule
	\end{tabular}
	}
        \caption{Comparison with state-of-the-art methods on SLR500 dataset. 3D-R50~\cite{qiu2017learning} is utilized for fusion with our method.}
	\label{slr500}
        \vspace{-1.5em}
\end{table}

\subsection{Ablation Study}

In this section, we conduct ablation studies to validate the effectiveness of our proposed approach and select proper hyper-parameters for our framework. 
For fair comparison, experiments are performed on the MSASL dataset and we report the top-1 accuracy under the per-instance and per-class metrics as the indicator.

\begin{table*}[t]
	\footnotesize
	\tabcolsep=26pt
	\begin{center}
		\resizebox{\textwidth}{!}{
		\begin{tabular}{lcccccc}
			\toprule
			\multirow{2}{*}[-0.13in]{Method}      &  \multicolumn{2}{c}{Total} &  \multicolumn{2}{c}{Confusing} &  \multicolumn{2}{c}{Normal}  \\
			\cmidrule(lr){2-3} \cmidrule(lr){4-5} \cmidrule(lr){6-7}
			& T-1  & T-5 & T-1 & T-5 & T-1 & T-5\\ \toprule
			\textbf{Pose-based}  & & & & & & \\
			ST-GCN~\cite{yan2018spatial} & 59.9 & 86.8 & 42.2 & 79.4 & 83.4 & 96.7 \\
 			SignBERT~\cite{hu2021signbert}  & 67.0 & \textbf{95.3} & 46.4  & \textbf{92.1} & 94.5 & 99.6 \\
			Ours & \textbf{68.5}  & 94.4 & \textbf{49.0} &  90.3 & \textbf{94.6} &  \textbf{99.7} \\ 
   \midrule
			\textbf{RGB-based}  & & & & & &  \\
			3D-R50~\cite{qiu2017learning}  & 62.1 & 82.9 & 43.1 & 72.4 & 87.4 & 97.0 \\
			DNF~\cite{cui2019deep} & 55.8 & 82.4 & 51.9 & 71.4 & 86.3 & 97.0 \\
			I3D~\cite{carreira2017quo}  & 64.4 & 88.0 & 47.3  & 81.8 & 87.1 & 97.3 \\
			TSM~\cite{lin2019tsm}   & 64.5 & 88.7 & 42.9 & 81.0 & 93.3 & 99.0 \\
			Slowfast~\cite{feichtenhofer2019slowfast} & 66.3 & 86.6 & 47.0 & 77.4 & 92.0 & 98.9 \\ 
			GLE-Net~\cite{hu2021global}  & 69.0  & 88.1 & 50.6 & 79.6 & 93.6 & 99.3 \\
			HMA~\cite{hu2021hand} & 64.7 & 91.0 & 42.3 & 84.8 & 94.6 & 99.3 \\
			Ours~(+R) & \textbf{79.2}  & \textbf{97.1}  &  \textbf{65.5} &  \textbf{95.0} & \textbf{97.5}  &  \textbf{99.9} \\ \bottomrule
		\end{tabular}
		}
	\vspace{-1em}
	\end{center}
        \caption{Comparison with state-of-the-art methods on NMFs-CSL dataset.  Our proposed method fused with another RGB-based method, 3D-R50~\cite{qiu2017learning}, is represented by Ours~(+R).}
	\label{NMFs-CSL}
	\vspace{-0.5cm}
\end{table*}

\begin{figure}[t!]
	\centering
	\includegraphics[width=1.0\linewidth]{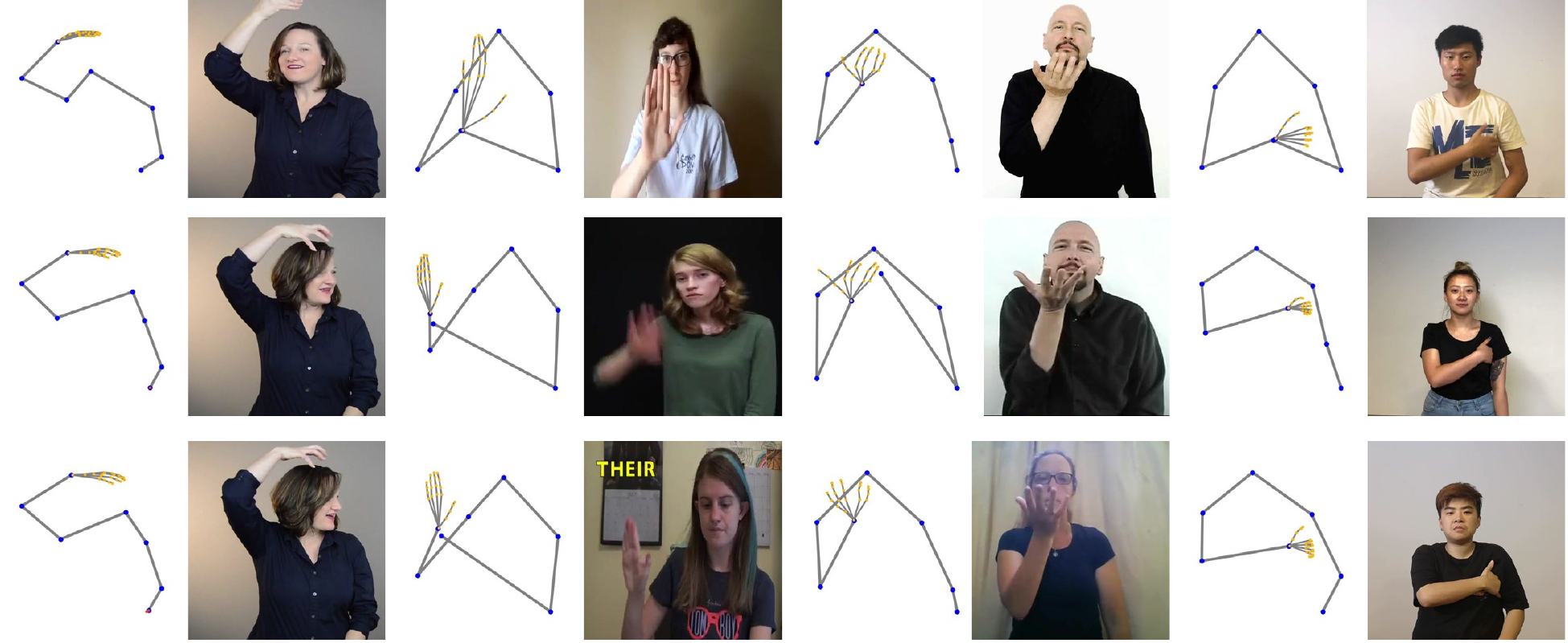}
	\caption{Qualitative results of our proposed coupling tokenization. Each two adjacent columns are organized as a group. In the same row of each group, the pose corresponds to the RGB image. Each group belongs to the same triplet unit label.}
	\label{fig:Visualization}
	\vspace{-1.5em}
\end{figure}

\noindent \textbf{Different Tokenizers.} 
In Table~\ref{quantizer}, we compare the impact of different tokenizers on the downstream SLR task. 
The K-Means method is a simple model-free clustering algorithm.
Different from the learnable tokenizer, K-Means iteratively transfers the 2D pose data to a series of clustering centers for hand and body, respectively.
Then we directly utilize the index of clustering centers as the pseudo labels.  
For fair comparison, the number of clustering centers is the same as the vocabulary size of our utilized codebook. 
VQ and Ours stand for separate and coupled vector quantizers, respectively. 
VQ learns the hand and body codebooks separately, while our proposed tokenizer jointly optimizes the hand and body codebooks.
It can be observed that our proposed tokenizer shows the best performance to validate our assumption. 
Since the body and hand are strongly correlated, our coupling tokenization utilizes this cue for better codebook learning.
Moreover, we illustrate the qualitative results of our proposed coupling tokenizer in Figure~\ref{fig:Visualization}.
Our tokenizer can successfully cluster similar pose triplet units under the unconstrained settings, \textit{i.e.,} filtering the disturbance from camera views, individual pose variance and inaccurate pose detection.

\begin{table}[t]
	\footnotesize
	\tabcolsep=10pt
	\begin{center}
	\resizebox{\linewidth}{!}{
		\begin{tabular}{l|cccccc}
			\toprule
			\multicolumn{1}{c|}{\multirow{2}{*}{Quantizer}} & \multicolumn{2}{c}{MSASL100} & \multicolumn{2}{c}{MSASL200} & \multicolumn{2}{c}{MSASL1000} \\
			\cmidrule(lr){2-3} \cmidrule(lr){4-5} \cmidrule(lr){6-7}
			& P-I & P-C & P-I & P-C & P-I & P-C  \\ \midrule
			K-Means   & 74.63 & 74.03 & 72.62 & 74.20 & 54.29 & 50.74 \\
			VQ  & 77.54 & 77.44  & 74.47 & 75.47 & 55.87 & 52.16 \\
			Ours & \textbf{80.98} & \textbf{81.24}  & \textbf{76.60} & \textbf{76.75} & \textbf{58.82} & \textbf{54.87} \\ \bottomrule        
		\end{tabular}
		}
        \vspace{-1em}
	\end{center}
         \caption{Comparison with different pose tokenizers on MSASL dataset. K-Means denotes a model-free clustering algorithm. VQ and Ours stand for separate and coupled vector tokenizers, respectively.}
         \label{quantizer}
        \vspace{-1.0em}
\end{table}

\begin{table}[t]
	\centering
	\tabcolsep=10pt
	\begin{center}
		\resizebox{\linewidth}{!}{
		\begin{tabular}{cc|cccccc}
			\toprule
			\multicolumn{2}{c|}{Mask} & \multicolumn{2}{c}{MSASL100} & \multicolumn{2}{c}{MSASL200} & \multicolumn{2}{c}{MSASL1000} \\
			\cmidrule(lr){3-4} \cmidrule(lr){5-6} \cmidrule(lr){7-8}
			Hand      & Body     & P-I & P-C &  P-I &  P-C &  P-I &  P-C\\ \midrule
			&            & 73.44 & 73.75 & 70.79 & 71.46 & 51.96 & 48.50 \\
			\checkmark &  & 76.75 & 77.22 & 71.37 & 72.07 & 54.60 & 51.63  \\
			& \checkmark & 68.82 & 68.82 & 67.11 & 67.84 & 47.72 & 43.87 \\
			\checkmark & \checkmark & \textbf{80.98} & \textbf{81.24} & \textbf{76.60} &\textbf{76.75}  & \textbf{58.82} &\textbf{54.87} \\ \bottomrule        
		\end{tabular}
	}
	\vspace{-1em}
	\end{center}
        \caption{Effect of the masking strategy on MSASL dataset. The first row denotes the baseline method without pre-training. ``Hand" and ``Body" denote the only masking on the hand and body part, respectively.}
	\label{mask}
        \vspace{-1em}
\end{table}

\noindent \textbf{Masking Strategy.} 
As shown in Table~\ref{mask}, we compare three different settings on our proposed MUM, \emph{i.e.,} only masking hand, only masking body and masking both parts~(hand and body).
The first row denotes the baseline without pre-training. 
It is observed that only masked hand setting shows better performance than only masked body setting due to the dominance of hand during sign language expression.
Our MUM adopts the third masking setting, which achieves the best performance.
Compared with the baseline, our proposed MUM achieves remarkable performance gain, \emph{i.e.,} 7.54\%, 5.81\% and 6.86\% for per-instance Top-1 accuracy improvement, respectively.

\noindent \textbf{Pre-Training Data Scale.} 
In Table~\ref{data_scale}, we investigate the effect of the pre-training data scale. 
The first row denotes that our proposed framework is the method without pre-training. It is clearly observed that the performance gradually increases with the increment in the proportion of pre-training data. The result demonstrates that our proposed method is applicable to the pre-training for large-scale data.

\begin{table}[t!]
	\footnotesize
	\tabcolsep=10pt
	\begin{center}
		\resizebox{\linewidth}{!}{
		\begin{tabular}{c|cccccc}
			\toprule
			\multicolumn{1}{c|}{\multirow{2}{*}{Percent}} & \multicolumn{2}{c}{MSASL100} & \multicolumn{2}{c}{MSASL200} & \multicolumn{2}{c}{MSASL1000} \\
			\cmidrule(lr){2-3} \cmidrule(lr){4-5} \cmidrule(lr){6-7}
			& P-I & P-C & P-I & P-C & P-I & P-C  \\ \midrule
			0\% & 73.44 & 73.75 & 70.79 & 71.46 & 51.96 & 48.50  \\
			25\% & 74.50 & 75.20 & 71.89 & 72.69 & 53.67 & 50.81 \\
			50\% & 77.54 & 78.06 & 73.43 & 75.03 & 54.69 & 51.47 \\
			75\% & 78.47 & 78.77 & 74.54 & 75.32 & 56.06 & 53.16  \\
			100\% & \textbf{80.98} & \textbf{81.24} & \textbf{76.60}  & \textbf{76.75} & \textbf{58.82}  & \textbf{54.87}   \\ \bottomrule  
		\end{tabular}
		}
	\vspace{-1.0em}
	\end{center}
        \caption{Effect of the data scale during pre-training on MSASL dataset. The ``Percent" denotes the proportion of pre-training data.}
	\label{data_scale}
        \vspace{-1.em}
\end{table}

\begin{table}[t!]
	\small
	\tabcolsep=4.5pt
	\begin{center}
		\resizebox{1.0\linewidth}{!}{
		\begin{tabular}{ll|cccccc}
			\toprule
			\multicolumn{1}{c}{\multirow{2}{*}{Mask.}} & \multicolumn{1}{c|}{\multirow{2}{*}{Obj.}} & \multicolumn{2}{c}{MSASL1000} & \multicolumn{2}{c}{WLASL2000} & \multicolumn{1}{c}{NMFs-CSL} & \multicolumn{1}{c}{\multirow{2}{*}{SLR500}}\\
			\cmidrule(lr){3-4} \cmidrule(lr){5-6} \cmidrule(lr){7-7}
			& & P-I & P-C & P-I & P-C & Total &  \\ \midrule
			- & - & 51.96 & 48.50 & 27.13 & 24.63 & 63.53  & 91.12 \\
			R-Mask & Regress. & 53.60 & 50.48 & 41.14 & 38.46 & 67.27 & 94.60 \\
			R-Mask & Token. & 53.95 & 51.11 & 42.91 & 40.42 & 64.12 & 93.75 \\
			MUM & Regress.  & 56.90 & 53.42 & 44.20 & 41.42 & 66.76 & 92.60 \\
			MUM & Token.  & \textbf{58.82} & \textbf{54.87} & \textbf{46.25} & \textbf{43.52} & \textbf{68.50} & \textbf{95.44} \\ \bottomrule     
		\end{tabular}
		}
	\end{center}
        \vspace{-1.0em}
        \caption{Comparison with different settings of self-supervised per-training. The first row denotes our framework without pre-training. ``Regress." and ``R-Mask'' are the objective function and masking strategy from SignBERT, respectively.
	``Token.'' and ``MUM'' denote those from ours.}
	\label{modeling}
        \vspace{-1.0em}
\end{table}

\noindent \textbf{Different Self-Supervised Pre-Training Settings.} 
The first row denotes the baseline method without pre-training.
We compare different settings of self-supervised pre-training. 
It is observed that our proposed mask strategy~\textit{MUM} and objective function~\textit{Token.} achieves better performance than other settings.
Notably, compared with the baseline, our method brings 19.12\% accuracy improvement on the WLASL2000 dataset.

\section{Conclusion}
In this paper, we propose a self-supervised pre-trainable framework namely BEST, which leverages the success of BERT with the specific design to the sign language domain.
Focusing on the main properties during sign language expression, we organize the hand and body movements as the pose triplet unit.
During pre-training, we propose the masked unit modeling~(MUM) pretext task to exploit the hierarchical context cues among internal and external triplet units.
To make the original BERT objective applicable, we attempt to bridge the semantic gap of pseudo label via coupling tokenization on the triplet unit.
After pre-training, we directly fine-tune our model with a prediction head on the downstream SLR task. 
Extensive experiments validate the effectiveness of our method, achieving new state-of-the-art performance on all four benchmarks.

\noindent \textbf{Acknowledgements.}
This work was supported by the National Natural Science Foundation of China under Contract U20A20183 and 62021001. 
It was also supported by GPU cluster built by MCC Lab of Information Science and Technology Institution, USTC.

\bibliography{aaai23}

\end{document}